# Different approaches for identifying important concepts in probabilistic biomedical text summarization

Milad Moradi, Nasser Ghadiri[1]

*Department of Electrical and Computer Engineering, Isfahan University of Technology,*

*Isfahan 84156-83111, Iran*

E-mail: `milad.moradi@ec.iut.ac.ir, nghadiri@cc.iut.ac.ir`

*Abstract*—Automatic text summarization tools help users in biomedical domain to acquire their intended information from various textual resources more efficiently. Some of the biomedical text summarization systems put the basis of their sentence selection approach on the frequency of concepts extracted from the input text. However, it seems that exploring other measures rather than the frequency for identifying the valuable content of the input document, and considering the correlations existing between concepts may be more useful for this type of summarization. In this paper, we describe a Bayesian summarizer for biomedical text documents. The Bayesian summarizer initially maps the input text to the Unified Medical Language System (UMLS) concepts, then it selects the important ones to be used as classification features. We introduce different feature selection approaches to identify the most important concepts of the text and to select the most informative content according to the distribution of these concepts. We show that with the use of an appropriate feature selection approach, the Bayesian biomedical summarizer can improve the performance of summarization. We perform extensive evaluations on a corpus of scientific papers in biomedical domain. The results show that the Bayesian summarizer outperforms the biomedical summarizers that rely on the frequency of concepts, the domain-independent and baseline methods based on the Recall-Oriented Understudy for Gisting Evaluation (ROUGE) metrics. Moreover, the results suggest that using the meaningfulness measure and considering the correlations of concepts in the feature selection step lead to a significant increase in the performance of summarization.

[1] *Corresponding author. Address: Department of Electrical and Computer Engineering, Isfahan University of Technology, Isfahan 84156-83111, Iran. Phone : +98-31-3391-9058, Fax: +98-31-3391-2450, Alternate email: nghadiri@gmail.com*

# 1. Introduction

Biomedical information available for researchers and clinicians is accessible from a variety of sources such as scientific literature databases, Electronic Health Record (EHR) systems, web documents, e-mailed reports and multimedia documents [1, 2]. The scientific literature provides a valuable source of information to researchers. It is widely used as a rich source for assessing the newcomers in a particular field, gathering information for constructing research hypotheses, and collecting information for interpretation of experimental results [3]. It is interesting to know that the US National Library of Medicine has indexed over 24 million citations from more than 5,500 biomedical journals in its MEDLINE bibliographic database[2]. However, a large amount of data cannot be effectively used to attain the desirable information in a limited time. The required information should be accessed easily at the right time, and in the most appropriate form [2]. For clinicians and researchers, efficiently seeking useful information from the ever-increasing body of knowledge and other resources is excessively time-consuming. Managing this information overload is shown to be a difficult task without the help of automatic tools.

Automatic text summarization is a promising approach to overcome the information overload problem, reducing the amount of text that must be read [4]. It can be used to obtain the gist efficiently on a topic of interest [1]. It helps the clinicians and researchers to save their time and effort required to seek information. Five reasons have been identified for producing summaries from full-text documents even when they provide abstracts [4]. The reasons include 1) there are variants of an ideal summary in addition to the abstract, 2) some content of the full-text may be missed in the abstract, 3) customized summaries are useful in question answering systems, 4) automatic summaries allow abstract services to scale the number of documents they can evaluate, and 5) assessing the quality of sentence selection methods can be helpful in development of multi-document summarization systems [4].

In recent years, various summarization methods have been proposed based on biomedical concepts [4-10]. They have improved the performance of biomedical text summarization, focusing on concepts extracted from the source text rather than terms. This concept-level analysis of text is performed with the help of biomedical knowledge sources, such as the Unified Medical Language System (UMLS). It has been shown that constructing a frequency distribution model from the concepts of the original text and following this model to create the summary yields better performance compared to traditional term-based methods [10]. Another successful biomedical summarization approach relies on concept chaining, identifying strong chains based on the frequency of their contained concepts, and sentence scoring according to the presence of the relevant concepts from the strong chains [9]. Regarding such summarization methods, some questions should be taken into account for summarizing a document. Do similar approaches, like probability distribution modeling of concepts, yield a desirable summarization performance for concept-based biomedical text summarization?

---

[2] http://www.nlm.nih.gov/databases/databases_medline.html

Should the summarizer consider all the concepts extracted from the source text? Are there any concepts which can be regarded as redundant and removed to increase the accuracy of the frequency or probability distribution model? Are there any criteria rather than the frequency to identify the important concepts? Can the model be more accurate by considering the correlations existing between concepts in the source text? In this paper, we address these questions, describing a Bayesian summarization method based on the probability distribution of concepts within the input document. We also introduce and evaluate different feature selection approaches to select the relevant concepts of the text and to use them as the classification features.

The Bayesian summarizer initially maps the input text to biomedical concepts contained in the UMLS [11], a well-known knowledge source in biomedical sciences maintained by the US National Library of Medicine. Then, it identifies important concepts and selects them as classification features. To this end, we discuss five different feature selection strategies based on various criteria and methods. The first strategy is the simplest one that considers all the extracted concepts. The second method discards the concepts which seem to be potentially redundant and unnecessary. The third method filters the concepts using the ranking method and according to the frequency of the concepts. The fourth method utilizes a meaningfulness measure defined by the Helmholtz principle [12] to identify the important concepts. The fifth method discovers the correlated concepts that represent the subtopics of the text by using frequent itemset mining [13], a well-known data mining technique. After the feature selection step, the summarizer represents each sentence as a vector of boolean features and specifies the value of features according to the occurrence of important concepts in the sentence. Afterwards, in the classification stage, it classifies the sentences into summary and non-summary classes using the naïve Bayes classification method [14]. The classifier estimates the posterior probability of classifying a sentence using the prior and likelihood probabilities of concepts. The summarizer selects the sentences that obtain the highest Posterior Odds Ratio (POR) values and puts them together to form the final summary.

To evaluate the performance of the proposed method, we conduct a set of experiments on a corpus of scientific papers from the biomedical domain and compare the results with other concept-based biomedical summarizers. We also evaluate the usefulness of the five different feature selection approaches to determine the competency of each one for this type of summarization. The results demonstrate that when the Bayesian summarizer uses the fourth and the fifth feature selection methods, it performs significantly better than the other frequency-based biomedical summarizers regarding the most commonly used Recall-Oriented Understudy for Gisting Evaluation (ROUGE) metrics [15].

The main contributions of this paper are:

- Using the naïve Bayes classifier in concept-based biomedical text summarization for classifying the sentences of a document based on the probability distribution of important concepts within the source text,

- Evaluating different feature selection approaches to identify the important concepts of a document and using them as classification features,
- Using different measure, i.e. the meaningfulness, rather than the frequency to determine the important concepts of a document and using them as classification features, and
- Discovering the correlated concepts of a document using itemset mining and using them as classification features.

The remainder of the paper is organized as follows. Section 2 gives an overview of text summarization, as well as a review of the related work in biomedical summarization. In Section 3, we introduce our biomedical summarization method based on a Bayesian classifier. We also define different feature selection approaches that can be utilized in our summarization process. Then, we describe the evaluation methodology in Section 4. The results of the preliminary experiments and the evaluations are presented in Section 5 and discussed in Section 6. Finally, Section 7 draws the conclusion and describes future lines of work.

## 2. Background and related work

### 2.1. Biomedical text summarization

Text summarization methods can be divided into *abstractive* and *extractive* approaches [1, 16]. An abstractive summarizer uses Natural Language Processing (NLP) methods to process and analyze the input text, then it infers and produces a new version. On the other hand, an extractive summarizer selects the most representative units (paragraphs, sentences, or phrases) from the original wording and puts them together into shorter form. Another classification of text summarization differentiates *single-document* and *multi-document* inputs [1, 2]. A single-document summarizer produces a summary which is the result of condensing only one document. In contrast, a multi-document summarizer gets a cluster of inputs and provides a single summary. Another classification of summarization methods is based on the requirements of users: *generic* vs. *user-oriented* (also known as query-focused summarizers) [1, 2, 17]. A general summary presents an overall implication of input document(s) without any specified preferences regarding content. While a user-oriented summary is biased towards a given query or some keywords to address a user's specific information requirement. Summarization systems can be *supervised* or *unsupervised* regarding whether they need training data [18]. A supervised system learns from labeled data to select the essential content of new documents, while an unsupervised system generates summaries for new documents without relying on any training data. In addition to the above categorizations, there are other types of summaries including *indicative*, *informative*, *multi-lingual*, *mono-lingual*, *cross-lingual*, *web-based*, *e-mail based*, *personalized*, *sentiment-based*, *survey*, and *update* summaries [18]. The Bayesian summarizer described in this paper is extractive, single-document, generic, and unsupervised.

In the biomedical field, various summarization methods have been proposed. These methods have been reviewed in a survey of early work [2] and in a systematic review of recently published research [1]. There have been some research works towards abstractive biomedical summarization. They could be regarded as tools for providing decision support data from MEDLINE citations [19], summarizing research related to the treatment of diseases [20], helping in evidence-based medical care [21], summarizing drug information [22], and multi-document summarization of MEDLINE citations [23, 24]. These methods mostly produce graphical summaries. On the other hand, the majority of extractive biomedical summarization systems focus on producing textual summaries. Extractive summarization methods have been widely studied in the biomedical domain for different tasks, such as summarizing clinical notes [25], developing clinical decision support tools for patient-specific recommendation and treatment [26, 27], and the summarization of EHRs [28].

Many biomedical summarizers utilize the UMLS knowledge source to map the input text to a wide range of biomedical and generic concepts. This mapping helps the systems to be domain-specific and act more accurately compared to traditional term-based methods. There are several knowledge sources such as MeSH, SNOMED, GO, OMIM, UWDA and NCBI Taxonomy widely used in knowledge-intensive data and text processing tasks in the biomedical domain. These knowledge sources along with over 100 controlled vocabularies, classification systems, and additional information sources have been unified into the UMLS. Plaza [29] performed an investigation on the impact of different knowledge sources on the performance of a summarization system. The evaluations showed that the quality of generated summaries was improved significantly with the use of an appropriate knowledge source. We make use of the UMLS concepts to incorporate the domain knowledge into the text modeling and the summarization process of our summarizer.

Some of the biomedical summarization methods employed graph representation along with the UMLS concepts for semantic modeling of the input text. Plaza et al. [5] proposed a graph-based approach to biomedical summarization. They used the UMLS concepts and the semantic relations between them to construct a semantic graph that is representative of the input document. Their system determined different topics within the text by applying a degree-based clustering algorithm on the semantic graph. Another work [30] performed the task of summarization based on a genetic graph-based clustering algorithm. Using the continuity of concept relations rather than the centroid method, it separated clusters and identified main topics. Menendez et al. [8] applied a combination of both genetic clustering and graph connectivity information to improve the performance of the previous semantic graph-based summarization systems. Compared to these approaches, our Bayesian summarization method utilizes a simpler modelling, representing the sentences of the input document as vectors of features. The features are important concepts within the input text.

Merging the domain knowledge and traditional methods, some domain-specific tools have been proposed for biomedical summarization. One of the studies [31] identified a set of medical cue terms and phrases and combined them with commonly used traditional features such as word frequency, sentence position, the similarity with the title of the article, and sentence length. The summarizer used the domain-specific and generic features for sentence scoring and summary generation. Sarkar et al. [32] proposed a supervised summarization method based on bagging and C4.5 decision tree as the base learner. They utilized the key terms in MeSH as a source of domain knowledge, as well as, other features including centroid overlap, first sentence overlap, sentence position, sentence length, and acronyms. In a hybrid summarization system [33], a classifier was utilized to learn and group sentences into six types of population, intervention, background, outcome, study, and other. The system also used a learning corpus to identify important UMLS concepts commonly appearing in summaries. Relative sentence position and sentence length were other features used by the summarizer. Another study [7] evaluated different positional approaches for sentence extraction in a semantic graph-based method for biomedical literature summarization. The study showed that sentences appearing in various sections of a biomedical article should be assigned different weights. We do not use any positional information in our summarization method. This allows the method to be applicable to input texts in which positional information may not be indicative of the importance and the informativeness of sentences.

Some of the biomedical summarizers use the frequency of UMLS concepts extracted from the source text as the basis of their summarization approach. BioChain method [9] pursued the lexical chaining idea [34], creating chains and putting each group of semantically related terms into a chain. BioChain used concepts rather than terms and put concepts belonging to the same semantic type into a chain. It computed the score of each chain using the frequency information of concepts, identified strong chains and concepts, and selected summary sentences according to the presence of strong concepts. FreqDist method [10] performed the task of sentence selection based on the frequency distribution of concepts within the source text. It initially created a frequency distribution model from the source text, also a summary frequency distribution model. Using an iterative sentence selection procedure, it selected a sentence that led to the closest alignment between the summary and original text frequency distributions in each iteration.

In contrast to BioChain and FreqDist, our method does not merely make use of the concept frequency. Employing the naïve Bayes classifier, it selects the sentences according to the probability distribution of concepts within the input text. It still benefits from the frequency information in the form of two coefficients that provide the classifier with additional knowledge. Compared to FreqDist, our method does not consider the distribution of all the extracted concepts. We evaluate different feature selection strategies to discard redundant and unnecessary concepts. Compared to BioChain, our method does not merely rely on the concept frequency to identify important concepts. We use another metric, namely the meaningfulness, in the form of a feature selection method that yields better summarization performance. Moreover, as one of the feature selection approaches, we extract correlated concepts and use each set of them as a classification feature. This correlation information provides the classifier with some additional knowledge to decide more accurately, leading to an increase in the performance of the summarizer.

In domain-independent text summarization, some methods have been proposed based on Bayesian approach. One of the basic methods [35] employed a set of features such as sentence length cut-off, fixed-phrase, paragraph, thematic word, and uppercase word to represent the sentences of a text document. It trained a Bayesian classifier on a training corpus and used the classifier to summarize new documents. BAYESUM [36], a supervised and query-focused multi-document summarizer built on Bayesian inference and language modeling techniques, represented documents and queries as probability distributions of words from a vocabulary. Estimating a sentence model for each sentence, it ranked sentences based on the language model and the similarity of sentences to the query model. Wang et al. [37] proposed a Bayesian sentence-based topic model for multi-document summarization. They employed a variational Bayesian algorithm to model the probability distribution of selecting sentences given topics and to estimate the model's parameters. Compared to these methods, our biomedical summarizer does not use any complicated topic modeling approaches and does not need any training data. It uses the prior probability of concepts to estimate the probability of selecting sentences for inclusion in the summary. We give the naïve Bayes classifier some additional knowledge in the form of two coefficients and different feature selection approaches.

To the authors' knowledge, no biomedical summarization method has been proposed so far based on the naïve Bayes classifier and estimating the probability of summary sentences by following the distribution of concepts within the source text. The rationale of our approach will be presented in Section 6.5 where it will be showed that for a corpus of 400 biomedical papers, the concepts within both the full-text papers and the ideal summaries (abstracts) follow the same distribution.

## 2.2. Bayesian classification

Our proposed summarization scheme consists of two main phases, preparation and classification. In the preparation phase, we perform concept extraction, feature selection, and sentence representation. In the classification phase, we utilize a Bayesian classifier to select sentences for the final summary. In the following, we give an overview of Bayesian classification [14].

In general, a Bayesian classifier is based on the Bayes theorem, defined by Eq. 1 below:

$$P(C|X) = \frac{P(X|C)P(C)}{P(X)} \tag{1}$$

where $C$ and $X$ are random variables. In classification tasks, they refer to observing class $C$ and instance $X$, respectively. $X$ is a vector containing the values of features. $P(C|X)$ is the posterior probability of observing class $C$ given instance $X$. In classification, it could be interpreted as the probability of instance $X$ being in class $C$, and is what the classifier tries to determine. $P(X|C)$ is the likelihood, which is the probability of observing instance $X$ given class $C$. It is computed from the training data. $P(C)$ and $P(X)$ are the prior probabilities of observing class $C$ and instance $X$, respectively. They measure how frequent the class $C$ and instance $X$ are within the training data. Using Eq. 1, the classifier can compute the probability of each class of target variable $C$ given instance $X$, and the most probable class, the class that maximizes $P(C|X)$, should be selected as the result of classification. This decision rule is known as Maximum A Posteriori (MAP). It is represented as follows:

$$C = \arg\max_{C_j} \frac{P(X|C = C_j)P(C = C_j)}{P(X)} \qquad (2)$$

where $C_j$ is the $j^{th}$ class (or value) of target variable $C$. in Eq. 2, the denominator is removed because it is constant and does not depend on $C_j$.

Making a conditional independence assumption, the naïve Bayes classifier reduces the number of probability values that must be estimated. It assumes that the probability of each value of feature $X_i$ is independent of the value of any other features, given the class variable $C_j$. Therefore, the naïve Bayes classifier finds the most probable class for the target variable by simplifying the joint probability calculation as follows:

$$C = \arg\max_{C_j} P(C = C_j) \prod_i P(X_i|C = C_j) \qquad (3)$$

The Posterior Odds Ratio (POR) is a well-known measure to assess the confidence of Bayesian classification. The POR shows a measure of the strength of evidence for a particular classification compared to other class values [38]. It is calculated as follows:

$$POR_i = \frac{P(C = C_j|X)}{P(C = C_k|X)} \qquad (4)$$

where $POR_i$ is the posterior odds ratio that measures the strength of evidence in favor of classifying the instance $X$ as class variable $C = C_j$ against classifying the instance $X$ as class variable $C = C_k$.

## 3. The Bayesian summarizer

Our Bayesian summarization method consists of five steps including (1) mapping text to biomedical concepts, (2) feature selection, (3) preparing sentences for classification, (4) sentence classification using naïve Bayes, and (5) summary generation. Fig. 1 illustrates the architecture of the Bayesian summarizer. In the following subsections, we give a detailed description of each step.

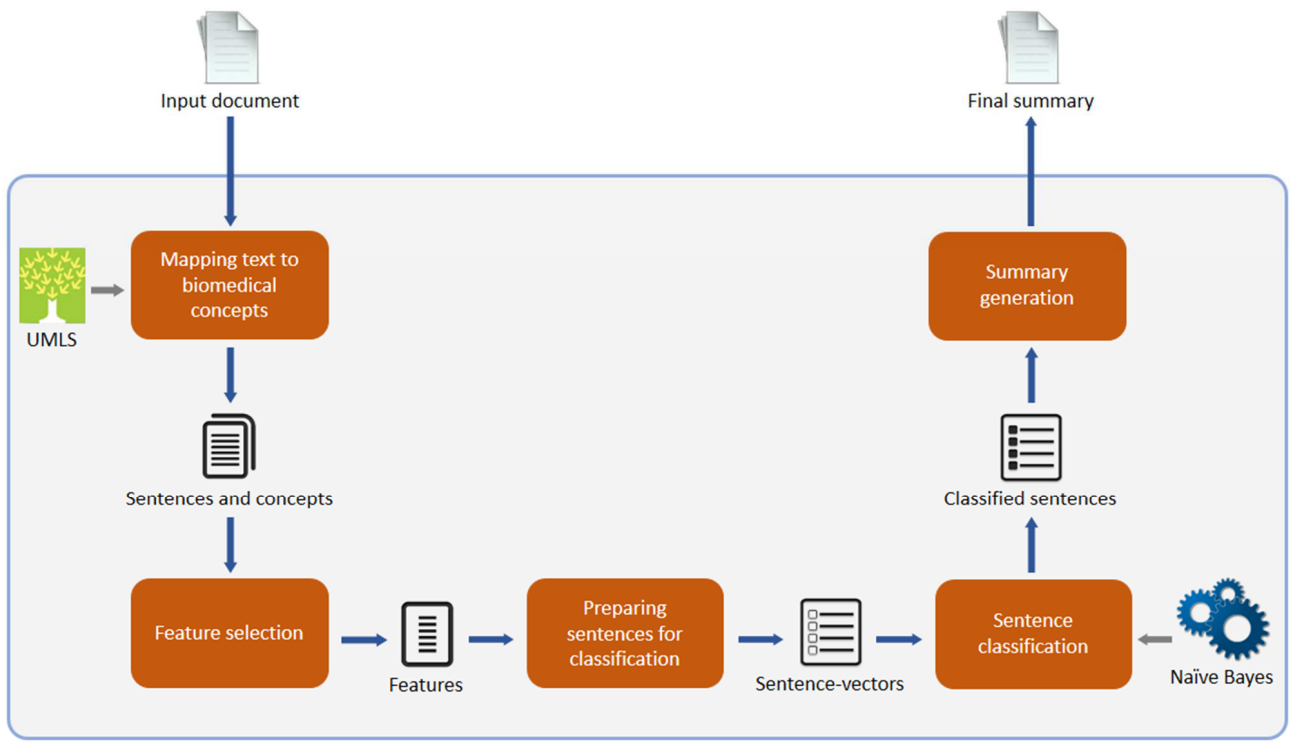

**Fig. 1.** The architecture of our proposed Bayesian biomedical text summarization method.

### 3.1. Mapping text to biomedical concepts

Firstly, the summarizer maps the input text to the concepts of the UMLS Metathesaurus. The Metathesaurus is a large, multi-lingual, and multi-purpose lexicon that contains millions of biomedical and health related concepts, their relationships and their synonymous names [39]. In addition to the Metathesaurus, the UMLS includes two main components, namely Specialist Lexicon and Semantic Network. The Specialist Lexicon is a lexicographic information database intended to use in NLP systems. It contains commonly occurring English words and biomedical vocabulary and records their syntactic, morphological and orthographic information [40]. The Semantic Network consists of a set of broad subject categories known as semantic types. These semantic types provide a categorization of all the concepts included in the Metathesaurus. It also contains a set of semantic relations between the semantic types [41].

For mapping biomedical text documents to the UMLS Metathesaurus concepts, the US National Library of Medicine has developed MetaMap program [42]. Using a knowledge-intensive approach based on NLP, computational linguistic and symbolic techniques, MetaMap identifies noun phrases in a text and extracts corresponding concepts. MetaMap may return multiple concepts when a noun phrase is mapped to more than one concept. In this situation, the summarizer selects all the mappings returned by MetaMap. It has been shown that the All Mappings strategy can be useful in concept-based biomedical text summarization [43]. MetaMap returns a semantic type along with each concept that determines the semantic category of the concept. As noted above, the semantic types are included in the UMLS Semantic Network. For the mapping step, we use the 2016 version of MetaMap program and the 2015AB UMLS release as the knowledge base. Fig. 2 shows a sample sentence and the concepts identified in the first step.

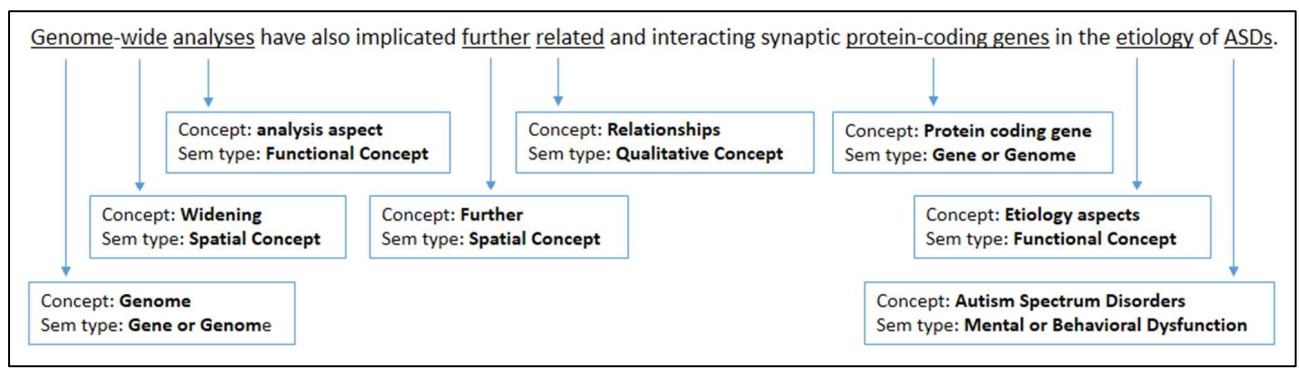

**Fig. 2.** A sample sentence and its identified concepts from the UMLS Metathesaurus.

### 3.2. Feature selection

In this step, the summarizer identifies important concepts within the input document. To this aim, we introduce five feature selection approaches. In the classification step, the summarizer uses the important concepts as classification features. We evaluate and discuss the impact of the feature selection approaches on the performance of the Bayesian summarizer in section 5 and 6.

In this subsection, we use a sample document[3] to present some examples of the feature selection strategies. The sample document is a scientific article about *Genetic overlap between autism, schizophrenia and bipolar disorders*. It contains 85 sentences.

#### 3.2.1. First approach: using all extracted concepts as features

The simplest approach to feature selection for the Bayesian summarizer is to consider all the distinct extracted concepts. It means that the summarizer decides about the summary and non-summary sentences considering the distribution of all the contained concepts regardless of whether they are important or not. For example, after concept extraction, the sample document contains 1042 concepts, 440 of which are distinct concepts that are used as the classification features. From these 440 distinct concepts, 267 concepts appear only one time in the document's sentences, and the three most frequent concepts appear in 30, 20, and 19 sentences.

We use the first feature selection approach as a baseline to assess the amount of improvement obtained by the other strategies. It also shows the impact of redundant and noisy features on the performance of the summarization method.

---

[3] Available at: http://genomemedicine.biomedcentral.com/articles/10.1186/gm102

*Manuscript*            10            *29 March 2017*

### 3.2.2. Second approach: filtering out generic features

There are some semantic types that their concepts can be discarded in the analysis of the input document. They are generic and broad concepts and almost frequently appear in majority of documents either in general English and biomedical texts. These semantic types have been identified empirically [5] and include *functional concept*, *qualitative concept*, *quantitative concept*, *temporal concept*, *spatial concept*, *mental process*, *language*, *idea or concept,* and *intellectual product*. In the second feature selection method, we remove concepts that belong to these semantic types and use remaining concepts as classification features. In this way, we remove a set of potentially redundant and misleading features, and we expect an improvement in the quality of produced summaries. Using this feature selection method, the summarizer discards the following concepts from the sentence represented in Fig. 2: *Widening*, *analysis aspect*, *Further*, *Relationships* and *Etiology aspects*.

As an example of the second approach, 234 distinct concepts remain after removing generic concepts from the sample document.

### 3.2.3. Third approach: filtering features by ranking

Filtering is a category of feature selection methods [44]. In filtering methods, features are scored based on some ranking criteria such as relevance, correlation, mutual information, and the frequency of occurrence [44, 45]. Then, the high-ranked features are selected according to a threshold or a predefined number of features. For the third feature selection strategy, we filter features by the frequency of their corresponding concepts in the text. After mapping the input text to the UMLS concepts and removing the potentially redundant features (similar to the second approach), the summarizer adds remaining features to a list named *feature_list*. Then, it ranks the features based on the frequency of corresponding concepts such that a higher rank is assigned to the feature that its corresponding concept appears more frequently. Finally, it filters the features of the *feature_list* using a threshold. It removes a feature if its frequency is less than the threshold value. In the following, we introduce three possible thresholds:

$$\theta_1 = \mu \quad (5)$$

$$\theta_2 = \mu + \sigma \quad (6)$$

$$\theta_3 = \mu + 2\sigma \quad (7)$$

where $\mu$ is the arithmetic mean of all concept frequencies in the *feature_list*, and $\sigma$ is the standard deviation of all concept frequencies in the *feature_list*.

In Section 5, we evaluate these three thresholds and use the best one as an optimum threshold for the third feature selection method.

As an example, when we use Eq. 5 as the threshold value, from the initial 440 distinct concepts of the sample document, 49 concepts are selected as features. When we use Eq. 6 and Eq. 7 as the threshold value, 25 and eight features are selected, respectively. Fig. 3 shows the identified important concepts for the sample document along with their semantic types and frequencies. In this example, we use Eq. 7 as the threshold value. The $\mu$, $\sigma$, and $\theta_3$ are equal to 2.435, 3.444, and 9.323 respectively. There are 234 concepts in the *feature_list*, eight of which are selected as features (Fig. 3).

- Schizophrenia, [*Mental or Behavioral Dysfunction*], Frequency: 30
- Bipolar Disorder, [*Mental or Behavioral Dysfunction*], Frequency: 20
- Autism Spectrum Disorders, [*Mental or Behavioral Dysfunction*], Frequency: 19
- Autistic Disorder, [*Mental or Behavioral Dysfunction*], Frequency: 17
- neuroligin, [*Molecular Function*], Frequency: 15
- Deletion Mutation, [*Genetic Function*], Frequency: 14
- Genes, [*Gene or Genome*], Frequency: 11
- NRXN1 gene, [*Gene or Genome*], Frequency: 11

**Fig. 3.** The identified important concepts by filtering (ranking and selecting based on a given threshold) for the sample document. The semantic types are represented in brackets. The features are sorted based on their frequencies.

### 3.2.4. Fourth approach: selecting meaningful features by the Helmholtz principle

The Helmholtz principle from the Gestalt theory of human perception defines a measure of meaningfulness for rapid change detection and keyword extraction in unstructured and textual data [12, 46]. In data mining context, the Helmholtz principle states that essential features and interesting events are observed in large deviations from randomness [47]. In text mining research, the Helmholtz principle has been used for document processing and keyword extraction [12], automatic text summarization [48], and supervised and unsupervised feature selection [49]. The primary study [12] dealt with words to extract the meaningful units of a text document, but we deal with concepts instead.

In the fourth method, the feature selection process is modeled as follows: Let *D* be the input document and *P* be a part of *D*. We consider *P* as a paragraph, but it can be any structural unit of the input document, such as a sentence or a page. After mapping the input document *D* to the UMLS concepts, we remove generic (potentially redundant) concepts, similar to the second approach, and add remaining concepts to the set *C*. For each concept *c* in the set *C*, we start to compute the meaningfulness measure by calculating the Number of False Alarms (NFA) in each *P*. If the concept *c* appears *m* times in the *P* and *k* times in the whole document *D*, then the NFA is calculated as follows:

$$NFA(c, P, D) = \binom{k}{m} \frac{1}{N^{m-1}} \qquad (8)$$

where *N* is equal to $[L/B]$ that *L* is the total number of concepts in the document *D*, and *B* is the total number of concepts in the paragraph *P*. In Eq. 8, $\binom{k}{m}$ is a binomial coefficient computed as follows:

$$\binom{k}{m} = \frac{k!}{m!\,(k-m)!} \tag{9}$$

To measure the meaningfulness value of concept *c* from *D* inside *P*, the following formula is used:

$$Meaning(c, P, D) = -\frac{1}{m} \log NFA(c, P, D) \tag{10}$$

Eventually, we construct a set *MeaningfulSet(ε)* of meaningful concepts. We add each concept *c* in *C* that its $Meaning(c, D)$ value is greater than *ε* to the *MeaningfulSet(ε)* where $Meaning(c, D)$ is the maximum of $Meaning(c, P, D)$ over all paragraphs *P*, and *ε* is a parameter that determines the level of meaningfulness. The summarizer selects the concepts included in the *MeaningfulSet(ε)* as classification features. In Section 5 and 6, we evaluate and discuss the optimum value of the parameter *ε*.

- Industrial machine, [*Manufactured Object*], Frequency: 2, Meaning: 1.445
- Promotion (action), [*Activity*], Frequency: 2, Meaning: 1.445
- Inhibition, [*Activity*], Frequency: 2, Meaning: 1.319
- Gene Knockout Techniques, [*Molecular Biology Research Technique*], Frequency: 2, Meaning: 1.319
- ethnic european, [*Population Group*], Frequency: 2, Meaning: 1.282
- Neurodevelopmental disorder, [*Mental or Behavioral Dysfunction*], Frequency: 2, Meaning: 1.282
- Research Activities, [*Research Activity*], Frequency: 5, Meaning: 1.282
- Mental disorders, [*Mental or Behavioral Dysfunction*], Frequency: 2, Meaning: 1.262
- genetic association, [*Occupation or Discipline*], Frequency: 2, Meaning: 1.242
- Synapses, [*Body Space or Junction*], Frequency: 5, Meaning: 0.991
- APBA2 gene, [*Gene or Genome*], Frequency: 3, Meaning: 0.867
- synaptogenesis, [*Organ or Tissue Function*], Frequency: 3, Meaning: 0.770
- Binding, [*Molecular Function*], Frequency: 7, Meaning: 0.741
- Procedure findings, [*Clinical Attribute*], Frequency: 6, Meaning: 0.711
- Genes, [*Gene or Genome*], Frequency: 11, Meaning: 0.587
- Reporting, [*Health Care Activity*], Frequency: 7, Meaning: 0.524
- Single Nucleotide Polymorphism, [*Nucleotide Sequence*], Frequency: 4, Meaning: 0.520
- Phenotype, [*Organism Attribute*], Frequency: 4, Meaning: 0.520
- neuroligin, [*Molecular Function*], Frequency: 15, Meaning: 0.509

**Fig. 4.** The identified meaningful concepts as the features by the Helmholtz principle (*ε=0.5*) for the sample document. The semantic types have been represented in brackets. The features have been sorted based on the descending order of their meaning measures.

For example, using this feature selection approach for the sample document, when *ε* is equal to 1, 0.5, 0, and −0.5, the number of selected features is 9, 19, 175, and 186 respectively. Fig. 4 shows the meaningful concepts of the sample document identified by the Helmholtz principle with a meaningfulness level of *ε=0.5*. As can be seen, there is no obvious relation between the meaning and frequency values. The meaning values depend on the appearing pattern of the concepts in the paragraphs and the whole document.

### 3.2.5. Fifth approach: extracting correlated features by itemset mining

In the four previous approaches, features are representative of single concepts. However, some correlations may be exist among multiple concepts. It means that some dependent concepts appear together in sentences and point to one of the subtopics of the text. In the fifth feature selection method, we utilize frequent itemset mining to extract correlated concepts and use each set of dependent concepts as a classification feature.

Frequent itemset mining is a data mining technique for finding items that appear together in a dataset [13]. It can be effectively used in our context to discover correlated concepts that frequently appear in the input text. For the fifth method, we utilize a well-known itemset mining algorithm, namely the Apriori [50]. Although the Apriori algorithm is usually used for association rule mining, we make use of its ability to extract frequent itemsets. This algorithm works with datasets structured in a transactional format. In a transactional dataset, there are several transactions each one contains some items. We can consider the input document as a transactional dataset such that the Apriori algorithm deals with each sentence and its contained concepts as a transaction and its items. Therefore, we perform itemset mining to discover frequent itemsets containing concepts that frequently appear in the input text.

Each itemset has a property, named *itemset support*, calculated by dividing the number of sentences that contain the itemset by the total number of sentences in the document. An itemset is said to be frequent if its support value is greater than or equal to a given minimum support threshold $\varphi$. A *k*-itemset is an itemset which contains *k* items. If all the subsets of a *k*-itemset are frequent, the itemset is said to be a frequent *k*-itemset. Given an input document, its extracted concepts, and a minimum support threshold $\varphi$, we perform the Apriori algorithm to discover correlated concepts. Each set of *k* correlated concepts discovered as a frequent *k*-itemset demonstrates a subtopic of the document. In Section 5 and 6, we evaluate and discuss the optimum value of the parameter $\varphi$.

- Autistic Disorder
- Bipolar Disorder
- Schizophrenia
Support: 0.105

- Autistic Disorder
- Bipolar Disorder
Support: 0.105

- Disease
Support: 0.094

- Genome-Wide Association Study
Support: 0.07

- neuroligin
- Binding
Support: 0.07

- Mental Retardation
Support: 0.07

- Bipolar Disorder
Support: 0.235

- Binding
Support: 0.082

- neuroligin
Support: 0.176

- Deletion Mutation
- NRXN1 gene
Support: 0.07

- Copy Number Polymorphism
Support: 0.105

- NRXN1 gene
Support: 0.129

- Alleles
Support: 0.094

- Encode
Support: 0.07

- Schizophrenia
- Autism Spectrum Disorders
Support: 0.07

- Procedure findings
Support: 0.07

- Autistic Disorder
Support: 0.2

- Genes
Support: 0.129

- Reporting
Support: 0.082

- Bipolar Disorder
- Schizophrenia
Support: 0.223

- Staphylococcal Protein A
Support: 0.07

- Scientific Study
Support: 0.082

- Proteins
Support: 0.082

- Study
Support: 0.094

- Autistic Disorder
- Schizophrenia
Support: 0.152

- Autism Spectrum Disorders
Support: 0.223

- Deletion Mutation
Support: 0.164

- Diagnosis
Support: 0.07

- Genome
Support: 0.105

- Schizophrenia
Support: 0.352

- Tryptophanase
Support: 0.094

- Persons
Support: 0.105

**Fig. 5.** The frequent itemsets extracted as the classification features from the sample document (minimum support threshold $\varphi=0.07$).

In the fifth feature selection strategy, after mapping the input document to the UMLS concepts, we remove generic concepts (similar to the second approach) and apply the Apriori algorithm. The summarizer uses the extracted frequent itemsets as classification features. For instance, when the minimum support threshold $\varphi$ is equal to 0.03, 0.07, and 0.1, the number of extracted frequent itemsets is 132, 32, and 15 respectively. Fig. 5 shows the extracted frequent itemsets extracted from the sample document. In this example, the value of minimum support threshold $\varphi$ is *0.07*. The Apriori algorithm returns a total number of 32 frequent itemsets. As can be seen, seven itemsets contain more than one concept. These itemsets convey the correlations that exist among some dependent concepts.

### 3.3. Preparing sentences for classification

After concept extraction and feature selection, the summarizer must represent the sentences of the input document in an appropriate format to be prepared for the classification step. It considers all the selected features as boolean such that for a given sentence, it sets the value of a feature to True if the corresponding concept appears in the sentence, otherwise it sets the value to False. Some features contain more than one concept when the summarizer uses the fifth feature selection method. For such features, the summarizer sets the value to True if all the corresponding concepts appear in the sentence.

After assigning feature values, the summarizer discards the sentences whose all feature values are False. We consider these sentences as unimportant and do not use them in the subsequent steps. For example, let the summarizer uses the third feature selection strategy and the threshold $\theta_3$ defined in Eq. 7 to filter the features for the sample document. Fig. 3 shows the selected features. The sample document consists of 85 sentences, of which 16 sentences do not contain any important concepts. The summarizer considers these 16 sentences as unimportant and discardeds them for the subsequent operations. Therefore, 69 sentences remain for preparation and classification. Note that we use this example hereafter to simplify the explanation of the remaining steps. In case of using the other feature selection approaches, the summarizer performs the preparation and classification stages likewise.

In this step, the summarizer creates a vector of features for each remaining sentence. It assigns a number to each vector according to the corresponding sentence number. For example, there are 69 vectors for the sample document, each one has eight features. Each feature corresponds to an important concept identified in the previous step. The summarizer assigns the values of features in the $i^{th}$ vector according to the presence and absence of important concepts in the $i^{th}$ sentence. For each vector, there is a target class or a target variable, named *Summary*, which is initially unknown. In the classification step, our Bayesian summarizer determines the value of the target class as *Yes* or *No* for all the vectors.

As an example, in the sample document, the $46^{th}$ sentence is "*Therefore, just as for NRXN1 deletions, it is apparent that these large CNVs confer risk of a range of neurodevelopmental phenotypes, including autism, mental retardation, and schizophrenia*". In this sentence, four important concepts can be seen, identified by the threshold $\theta_3$ and represented in Fig. 3, including *Autistic Disorder*, *Schizophrenia*, *Deletion Mutation* and *NRXN1 gene*. Hence, in the *46th* vector, the values of these four features are *True,* and the values of the other features are *False*. Fig. 6 shows the *46th* vector in the sample document. The summarizer assigns the feature values for all the vectors just in the same way as the above example.

After this step, we have a collection of vectors and their feature values. Their class variables are unknown, and the summarizer must classify them as summary sentences (*Yes*) or non-summary sentences (*No*). Every document has its particular set of concepts and the features of each text are different from others.

| Number | $f_1$ | $f_2$ | $f_3$ | $f_4$ | $f_5$ | $f_6$ | $f_7$ | $f_8$ | Summary |
|---|---|---|---|---|---|---|---|---|---|
| 46 | True | False | False | True | False | True | False | True | - |

$f_1$: Schizophrenia  
$f_2$: Bipolar Disorder  
$f_3$: Autism Spectrum Disorders  
$f_4$: Autistic Disorder  
$f_5$: neuroligin  
$f_6$: Deletion Mutation  
$f_7$: Genes  
$f_8$: NRXN1 gene

**Fig. 6.** The *46th* sentence-vector, corresponding to the *46th* sentence in the sample document. The *Summary* class variable is initially unknown.

### 3.4. Sentence classification

The compression rate is a parameter in summarization systems which determines the percentage of the input text that must be extracted as the final summary. The summarizer does not initially know about the $P(C = C_j)$ in Eq. 3, the prior probability of class variable values. However, it knows what percentage of sentences that must be selected as summary, and it can estimates the $P(Summary = Yes)$ and $P(Summary = No)$. For instance, the total number of sentences in the sample document is 85. Suppose the compression rate is 0.3. It means that 30% of the text (about 26 sentences) must be selected for the final summary. In the preparation step, the summarizer discarded the sentences that did not include any important concepts, and 69 vectors remained for the classification step. The summarizer does not know which of these 69 vectors has *Yes* value for the *Summary* class variable, but it knows 26 vectors must have the *Yes* value. Hence, for this example, the $P(Summary = Yes)$ is equal to 26 / 69 = 0.377 and $P(Summary = No)$ is equal to 43 / 69 = 0.623. Generally, the $P(Summary = Yes)$ is equal to the number of sentences that must be selected for the final summary (according to the compression rate) divided by the total number of remaining sentences for the classification step.

The Bayesian summarizer follows an assumption about the $P(x_i|C = C_j)$ in Eq. 3 that simplifies the estimation of the likelihood probabilities. The summarizer assumes that the summary can convey an informative content if it follows the distribution of important concepts within the input document. With regard to this assumption, the summarizer can estimate the likelihood probabilities, i.e. the probability of observing an important concept given class *Yes* or *No*. For example, the concept *Schizophrenia* appears in 30 sentences within the sample document, and its distribution within all the vectors is equal to 30 / 69 = 0.435. Therefore, the probability of observing concept *Schizophrenia* given class *Yes*, i.e. the $P(Schizophernia = True|Summary = Yes)$, would be equal to 0.435. Likewise, the probability of not observing concept *Schizophrenia* given class *Yes*, i.e. $P(Schizophernia = False|Summary = Yes)$, would be equal to 1 − 0.435 = 0.565. Likewise, the summarizer estimates the likelihood probabilities of observing and not observing a concept given class value *No*.

In this step, the summarizer can estimate the posterior probability of class values given a vector. If the summarizer selects the value that maximizes the posterior probability of *Summary* variable given $i^{th}$ vector (similar to Eq. 3) the number of sentences classified as *Yes* may be less than the number of sentences that must be selected for the summary. This comes true, because in a vector the number of features having the True value is often less than the number of features having the False value. Therefore, the summarizer should decide about the *Summary* class values in a different way compared with Eq. 3. The summarizer estimates the posterior probability of class values for each vector and assesses the strength of evidence for class value *Yes* using the POR measure.

We incorporate two coefficients into the estimation of posterior probabilities to discriminate between the presence and absence of more important and less important concepts. In our context, the presence of important concepts (the *True* value for the features) and their prior probabilities are contributing factors in the selection a sentence for the summary. On the other hand, when the suumarizer uses the Bayes rule, it does not discriminate between the presence and the absence of more important (high-frequent) and less relevant (low-frequent) concepts. In this case, the summarizer decides based on the highest probable values of features. As we show in Section 5.1.2, for the majority of documents, even the high-frequent concepts appear in less than 50% of the sentences of a document. This shows that for the majority of documents, the most probable value for all features is *False*. Therefore, for estimating the posterior probabilities of the class value *Yes* and *No*, the summarizer should take into account this issue. We address this problem employing two coefficients in estimation of the posterior probability of class values. The coefficients increase and decrease the impact of important concepts on the posterior probability of class values based on the frequency of concepts and whether they occur in a sentence or not. We evaluate and discuss the impact of using these coefficients on the accuracy of the Bayesian summarizer in Section 5 and 6.

### 3.4.1. The probability of inclusion in the summary

To estimate the posterior probabilities, firstly, the summarizer estimates the posterior probability of class value *Yes* given $i^{th}$ vector by rewriting Eq. 3, as follows:

$$P(Summary = Yes|V_i) = P(Summary = Yes) \prod_k P(f_{ik}|Summary = Yes)(C_1) \qquad (11)$$

where $V_i$ is the $i^{th}$ vector, $P(Summary = Yes|V_i)$ is the posterior probability of classifying the $i^{th}$ vector as *Yes* given $V_i$, $f_{ik}$ is the $k^{th}$ feature in the $i^{th}$ vector, and $P(f_{ik}|Summary = Yes)$ is the likelihood probability, i.e. the probability of observing $f_{ik} = True$ or $f_{ik} = False$ given class variable $Summary = Yes$. The value of the coefficient $C_1$ depends on whether the $f_{ik}$ is *True* or *False* and is specified as follows:

$$C_1 = \begin{cases} freq_{ik} & if\ f_{ik} = True \\ \dfrac{1}{freq_{ik}} & if\ f_{ik} = False \end{cases} \qquad (12)$$

where $freq_{ik}$ is the frequency of the concept corresponding to the $f_{ik}$. Depending on whether the value of $f_{ik}$ is *True* or *False*, the coefficient $C_1$ affects the $P(f_{ik}|Summary = Yes)$ in two ways:

1. When the $f_{ik}$ is *True*, the frequency of corresponding concept is multiplied by the $P(f_{ik}|Summary = Yes)$. Thus, the values of the $P(f_{ik}|Summary = Yes)$ and $P(Summary = Yes|SV_i)$ increase. Consequently, the presence of more frequent concepts increases the chance of selecting a sentence for the summary.

2. When the $f_{ik}$ is *False*, the inverted frequency of corresponding concept is multiplied by the $P(f_{ik}|Summary = Yes)$, and as a result, the values of the $P(f_{ik}|Summary = Yes)$ and $P(Summary = Yes|SV_i)$ decrease. In this case, a higher frequency decreases the $P(f_{ik}|Summary = Yes)$ with a higher rate. Hence, the absence of more frequent concepts decreases the chance of selecting a sentence for the summary.

### 3.4.2. The probability of exclusion from the summary

After estimating the probability of inclusion in the summary, the summarizer estimates the posterior probability of class value *No* given $i^{th}$ vector as follows:

$$P(Summary = No|V_i) = P(Summary = No) \prod_k P(f_{ik}|Summary = No)(C_2) \qquad (13)$$

where $V_i$ is the $i^{th}$ vector, $P(Summary = No|V_i)$ is the posterior probability of classifying the $i^{th}$ vector as *No* given $V_i$, $f_{ik}$ is the $k^{th}$ feature in the $i^{th}$ vector, and $P(f_{ik}|Summary = No)$ is the likelihood probability, i.e. the probability of observing $f_{ik} = True$ or $f_{ik} = False$, given class variable $Summary = No$. The value of the coefficient $C_2$ depends on whether the $f_{ik}$ is *True* or *False* and is specified as follows:

$$C_2 = \begin{cases} \dfrac{1}{freq_{ik}} & if\ f_{ik} = True \\ freq_{ik} & if\ f_{ik} = False \end{cases} \quad (14)$$

where $freq_{ik}$ is the frequency of the concept corresponding to the $f_{ik}$. Similar to the $C_1$, depending on whether the value of $f_{ik}$ is *True* or *False*, the $C_2$ affects the $P(f_{ik}|Summary = No)$ in two ways:

1. When the $f_{ik}$ is *True*, the inverted frequency of corresponding concept is multiplied by the $P(f_{ik}|Summary = No)$, and as a result, the values of the $P(f_{ik}|Summary = No)$ and $P(Summary = No|SV_i)$ decrease. In this case, a higher frequency decreases the $P(f_{ik}|Summary = No)$ with a higher rate. Hence, the presence of more frequent concepts decreases the probability of not selecting a sentence for the summary.

2. When the $f_{ik}$ is *False*, the frequency of corresponding concept is multiplied by the $P(f_{ik}|Summary = No)$. Thus, the values of the $P(f_{ik}|Summary = No)$ and $P(Summary = No|SV_i)$ increase. Consequently, the absence of more frequent concepts increases the probability of not selecting a sentence for the summary.

After estimating the probability of classifying each vector as *Yes* and *No*, the summarizer needs to decide which sentences should be selected for inclusion in the final summary. As mentioned earlier, if the classifier selects for each vector the class value which maximizes the posterior probability of *Summary* class variable, the number of sentences classified as *Yes* may be less than the number of sentences required for the summary. Therefore, we employ the POR measure, early explained in Section 2.2, to classify the vectors. For the $i^{th}$ vector, the summarizer computes the value of the POR by rewriting Eq. 4, as follows:

$$POR_i = \frac{P(Summary = Yes|SV_i)}{P(Summary = No|SV_i)} \quad (15)$$

where $POR_i$ is the posterior odds ratio of the $i^{th}$ vector. The values of the $P(Summary = Yes|SV_i)$ and $P(Summary = No|SV_i)$ are the posterior probabilities of classifying the $i^{th}$ vector as *Yes* and *No* given $SV_i$, which the summarizer estimated earlier.

The POR demonstrates a measure of the strength of evidence for a particular class value. Therefore, the greater $POR_i$ for a vector, the higher strength of evidence in favor of classifying the vector as *Yes*.

After calculating the POR value for all the vectors, the summarizer can decide which sentences should be selected for the summary. It sorts the vectors in descending order of their POR values and assigns the top-ranked $N$ vectors to the class *Yes* where $N$ is the number of sentences which must be selected to make the summary specified by the compression rate. The summarizer assigns the remaining vectors to the class *No*.

In the following, we explain a redundancy reduction method that the summarizer can use to decrease redundant information in the summary.

### 3.4.3. The redundancy reduction method

The problem of redundancy in text summarization concerns the same repeated information conveyed by multiple sentences in a summary. Compared to multi-document summarization, it is less probable to find redundant information in a summary produced for a single document [51]. However, redundancy removal approaches can also be useful in single-document summarization [10]. Maximal Marginal Relevance (MMR) [52] is a well-known method for removing redundancy, especially in query-focused summarization. It computes cosine similarities between sentences and a query, also between sentences and already selected sentences. Then, it assigns a marginal relevance to each sentence and adds the sentence with the maximum marginal relevance to the summary. The MMR computes a linear combination of two functions, i.e. relevance and novelty. The relevance function needs a query to assess the relatedness of sentences. Since the Bayesian summarizer does not use any query for summarization, the MMR approach is not applicable to our method. Hence, we propose a redundancy reduction method based on our context by gradually updating the probabilities to decrease the chance of high-probable concepts and increase the chance of less-probable concepts for inclusion in the summary.

We employ an iterative method aimed at reducing the redundancy that can emerge in the summary due to the large prior probability of high-frequency concepts. When the possible redundancy does not matter to the summarizer, as explained earlier, the summarizer estimates the prior, likelihood and posterior probabilities. Then, it computes the POR values for all the sentences and ranks them based on their POR values. Finally, it Assigns the top *N* sentences to the class value *Yes*. On the other hand, when the summarizer employs the redundancy reduction method, it performs the sentence selection process differently. In this case, when it computes the POR values, it assigns only the sentence having the highest POR value to the class value *Yes*. Next, it estimates the prior, likelihood, and posterior probabilities again, but it does not consider the previously selected sentences and their concepts in the subsequent estimations. Accordingly, it reduces the probability of observing the high-frequency concepts included in the sentences already selected. Moreover, the summarizer increases the chance of observing the low-frequency concepts in the summary. In the subsequent iterations, it computes the POR values based on new probabilities. It repeatedly estimates the probabilities without considering the sentences already selected for the summary and selects the sentence with the maximum *POR* value until the number of summary sentences (assigned to the class value *Yes*) reaches *N*. Finally, it assigns the remaining sentences to the class value *No*.

In Section 5 and 6, we evaluate and discuss the efficiency of the redundancy reduction method.

### 3.5. Summary generation

In the previous step, the summarizer assigned the sentences to the class values *Yes* and *No*. In the summary generation step, it adds the sentences of the class value *Yes* to the summary. It arranges the summary sentences in the same order as they appear in the primary document. Finally, it adds the figures and tables in the main document referred to in the summary. Fig. 7 shows the summary of the sample document produced by the Bayesian summarizer. In this example, for brevity reasons, the compression rate is 10%. It means that the size of the summary must be 10% of the input document.

> Psychiatric 'disorders' such as autism, schizophrenia and bipolar disorder are therefore effectively groups of symptoms making up syndromes that define groups of patients who show broadly similar outcomes and who respond similarly to treatment.
> Although the available data provide relatively strong evidence that disruption of the Neurexin-1 locus (NRXN1) is a risk factor for schizophrenia and ASDs, evidence in relation to bipolar disorder is lacking.
> In contrast, there is evidence that the global burden of duplications or deletions in bipolar disorder is substantially less than for schizophrenia and ASDs.
> Therefore, just as for NRXN1 deletions, it is apparent that these large CNVs confer risk of a range of neurodevelopmental phenotypes, including autism, mental retardation and schizophrenia.
> So far there have been no systematic comparisons of GWAS data for ASDs with those from schizophrenia or bipolar disorder.
> Furthermore, there have been recent reports of association for common alleles at several GABA receptor genes in a subtype of bipolar disorder and schizophrenia, which implicate loci also reported as associated with ASDs.
> Such biological roles fit with hypotheses of the etiology of autism and schizophrenia in which a neurodevelopmental insult and adult imbalance in excitatory and inhibitory neurotransmission occur in the absence of overt macro-pathology. SHANK3 is implicated in autism by several lines of evidence and functions as a post-synaptic scaffolding protein that binds indirectly to neuroligins, forming a potentially functional circuit of neurexin-neuroligin-Shank that is dysregulated in ASDs.
> Therefore, the evidence from ASDs, schizophrenia and bipolar disorder suggests a convergence on specific processes involved in the development and regulation of synaptic transmission.
> We have based this conclusion on the fact that several rare CNVs, including deletions of NRXN1, are associated with mental retardation, autism and schizophrenia, and on the overlap in common risk alleles seen between schizophrenia and bipolar disorder.

**Fig. 7.** The summary of the sample document generated by the Bayesian summarizer (compression rate=10%).

## 4. Evaluation method

### 4.1. Evaluation corpus

The most common method of evaluating summaries generated by an automatic summarizer is to compare them against manually generated summaries, known as model or reference summaries. In such evaluation method, we measure the similarity between the content of system and model summaries. The more content shared between system and model summaries, the better the performance of the summarization system. Obtaining manually generated summaries is a challenging and time-consuming task, because they have to be provided by human experts. Moreover, human-generated model summaries are highly subjective. To the authors' knowledge, there is no corpus of model summaries for single-document biomedical text summarization. However, most scientific papers have an abstract which can be considered as the model summary for evaluation [5].

To compare our Bayesian biomedical summarization method against other summarizers, we use a collection of 400 biomedical scientific papers randomly selected from the BioMed Central's corpus for text mining research[4]. The size of evaluation corpus is large enough to allow the results of the assessment to be significant [53]. We use the abstracts of the papers as the model summaries to evaluate the performance of the system-generated summaries. We perform our preliminary experiments using a separate development corpus containing 100 papers randomly selected from the BioMed Central's corpus.

### 4.2. Evaluation metrics

A common feature which we assess in the performance evaluation of text summarization systems is the informativeness. It is a feature for representing how much information from the original text is provided by the summary [54]. In this paper, we use the ROUGE package [15] to evaluate the performance of the Bayesian summarizer in terms of the informative content of summaries. The ROUGE package compares a system-generated summary with one or more model summaries, estimates the shared content between them by calculating the proportion of shared *n*-grams, and produces different scores in terms of different metrics. The ROUGE metrics produce a score between 0 and 1. The higher scores for a system summary, the greater content overlap between the system and model summaries. In our evaluations, we use the following ROUGE metrics:

- ROUGE-1 (R-1). It computes the number of shared unigrams (1-grams) between the system and model summaries.
- ROUGE-2 (R-2). It computes the number of shared bigrams (2-grams) between the system and model summaries.
- ROUGE-W-1.2 (R-W-1.2). It computes the union of the longest common subsequences between the system and model summaries. It takes into account the presence of consecutive matches.
- ROUGE-SU4 (R-SU4). It computes the overlap of skip-bigrams (pairs of words having intervening word gaps) between the system and model summaries. It allows a skip distance of four between bigrams.

In spite of their simplicity, the ROUGE metrics have shown a high degree of correlation with human judges [15].

### 4.3. Preliminary experiments and parameterization

We introduce five feature selection approaches in Section 3.2. The first approach selects the classification features by simply considering all the extracted concepts. The second approach tries to reduce the number of features by filtering out the generic features that seem to be potentially redundant. In the third approach, we rank the features based on the frequency of corresponding concepts and use a threshold as a filtering criterion. We introduce three possible threshold values in Section 3.2.3 based on the average and the standard deviation of the frequency of concepts. In feature selection experiments, we evaluate these three values to specify the optimum threshold for this type of filtering. In the fourth feature selection approach, we measure a meaningfulness value for each concept using the Helmholtz principle. We use a parameter $\varepsilon$ that determines the level of meaningfulness for the concepts selected as classification features. We perform a set of experiments to tune the parameter $\varepsilon$. In the fifth approach, we utilize an itemset mining method to extract correlated concepts in the form of frequent itemsets and use them as classification features. The itemset mining extracts frequent itemsets according to a minimum support threshold $\varphi$. We tune the parameter $\varphi$ performing a set of preliminary experiments.

---

[4] http://old.biomedcentral.com/about/datamining

In the other set of preliminary experiments, we assess the impact of the coefficients $C_1$ and $C_2$, introduced in Section 3.4, on the performance of our summarization method. We evaluate the quality of the produced summaries in two situations, the presence and the absence of the coefficients.

In Section 3.4.3, we introduce a redundancy reduction method to decrease the potential redundancy in the summary. We conduct another set of experiments to investigate the impact of the redundancy reduction method on the performance of the Bayesian summarizer. It seems that the redundancy reduction strategy may achieve more percent of improvement under smaller compression rates. We also assess the percentage of improvement under smaller and larger compression rates than 30%.

### 4.4. Comparison with other summarization methods

We compare the Bayesian summarizer with six summarization methods and two baselines. Three methods of comparison systems are biomedical summarizers, i.e. *FreqDist* [10], *BioChain* [9], and *ChainFreq* [4]. We implement these three summarizers as explained in their original papers. Two comparison methods are domain-independent and term-based, i.e. *SUMMA* [55] and *SweSum* [56]. One of the methods, *Microsoft AutoSummarize*, is a commercial application. The two baseline methods are *Lead baseline* and *Random baseline*. The size of the summaries generated by all the summarizers is 30% of the original documents. The choice of 30% as the compression rate is based on a well-accepted de facto standard that says the size of a summary should be between 15% and 35% of the original text [57]. In the following, we give a brief description of the competitor methods.

**FreqDist** [10] is a biomedical summarization method which uses concept frequency distribution to identify important sentences. It initially maps the input text to the UMLS concepts. Then, it creates an empty summary frequency distribution and a source text frequency distribution model counting the concepts. Afterwards, using an iterative sentence selection process, FreqDist creates a candidate summary and compares the frequency distribution of the candidate summary with the distribution of the source text. The method evaluates the sentences based on how much they align the frequency distribution of the summary to the original text. It

*Manuscript*            23            *29 March 2017*

selects a sentence in each iteration and adds it to the summary such that the two frequency distributions to be aligned as closely as possible. The original study has compared five similarity functions to find the best one for evaluating the similarity of frequency distributions. We implement and compare FreqDist method with the Dice's coefficient as it has reported the highest ROUGE scores in the original study.

**BioChain** [9] is a biomedical summarizer based on the lexical chaining idea. It extracts the UMLS concepts from the input document, considers the semantic types as the head of chains, and puts the concepts of the same semntic type in the same chain. BioChain selects the strong chains based on the core concepts and their frequency, identifies the strong concepts of each strong chain, and uses the strong concepts to score the sentences. Finally, it extracts the high-scoring sentences and generates the final summary.

**ChainFreq** [4] is a hybrid summarizer which makes use of both FreqDist and BioChain methods. It uses BioChain to identify important sentences containing strong concepts. Then, it sends the candidate sentences to FreqDist to reduce the redundancy and to select the subset of sentences which aligns the summary frequency distribution to the source text. In the original study, two variants of BioChain have been evaluated for ChainFreq. The first variant uses all the concepts of strong chains to score the sentences, while the second one uses the most frequent concept of each strong chain. In the evaluations, the first method has obtained higher ROUGE scores. Accordingly, we implement the first method as a part of ChainFreq for our evaluations. We also implement FreqDist using the Dice's coefficient.

In addition to the above biomedical summarizers, we use three domain-independent comparison methods in the evaluations to assess the performance of our method against traditional term-based approaches. We give a description of these methods in the following.

**SUMMA** [55] is a summarizer which uses generic and statistical features to score the sentences of an input document. The features that we use for the evaluations include the frequency of terms within the sentence, the position of the sentence within the document, the similarity between the sentence and the first sentence of the document, and the similarity between the sentence and the title.

**SweSum** [56] is an online and multi-lingual summarizer based on generic features. For the evaluations, we use the following set of features: presence of the sentence in the first line of the text, presence of numerical values in the sentence, and presence of keywords in the sentence. We also set the type of text feature to 'Academic'.

**Microsoft AutoSummarize** is a feature of the Microsoft Word software[5]. This summarizer performs based on a word frequency algorithm. It assigns a score to each sentence of a document according to the frequency of words contained in the sentence.

---

[5] Microsoft Word 2007, Microsoft Corporation

Our two baselines for the evaluations are **Lead baseline** that returns the first *N* sentences of the input document as the summary, and **Random baseline** that randomly selects *N* sentences from the document and generates a summary.

## 5. Results

### 5.1. Preliminary experiments

In this subsection, we first present the results of parameterization and the preliminary experiments which specifies the best settings for the feature selection approaches. Then, we present the results of experiments conducted to assess the impact of the coefficients and the redundancy reduction method on the performance of the Bayesian summarizer. For brevity reasons, we only report the R-2 and R-SU4 scores for the preliminary experiment results.

#### 5.1.1. Feature selection

As explained in Section 4.3, we conduct a set of preliminary experiments to tune the parameters and find the best settings of the feature selection methods. We introduce three possible threshold values for the third method which uses a ranking and filtering strategy. Table 1 shows the ROUGE scores obtained by the Bayesian summarizer using the three thresholds. The summarizer obtains the highest scores when it uses the threshold $\theta_3$. We use this threshold as the optimal value in the subsequent experiments.

**Table 1.** ROUGE scores obtained by the Bayesian summarizer using the third feature selection approach and three threshold values. The best result for each ROUGE score is shown in bold type.

|  | *ROUGE-2* | *ROUGE-SU4* |
|---|---|---|
| $\theta_1 = Avg(freq)$ | 0.3188 | 0.3769 |
| $\theta_2 = Avg(freq) + Std\_dev(freq)$ | 0.3242 | 0.3828 |
| $\theta_3 = Avg(freq) + (2 \times Std\_dev(freq))$ | **0.3346** | **0.3932** |

In the fourth feature selection approach, we use the Helmholtz principle to identify meaningful concepts. In this method, there is a parameter $\varepsilon$ which specifies the meaningfulness threshold. Fig. 8 shows The ROUGE scores obtained by the Bayesian summarizer using the fourth feature selection method and different values of the theshold $\varepsilon$ between [−1.5, 0.6]. The threshold values of −1.3 and −1.2 report the best scores (R-2: **0.3411** and R-SU4: **0.3980**). We set the optimal value of this parameter to −1.2 in the subsequent experiments. Fig. 9 shows the average number of meaningful concepts selected as classification features for the different values of the threshold $\varepsilon$ in the given range.

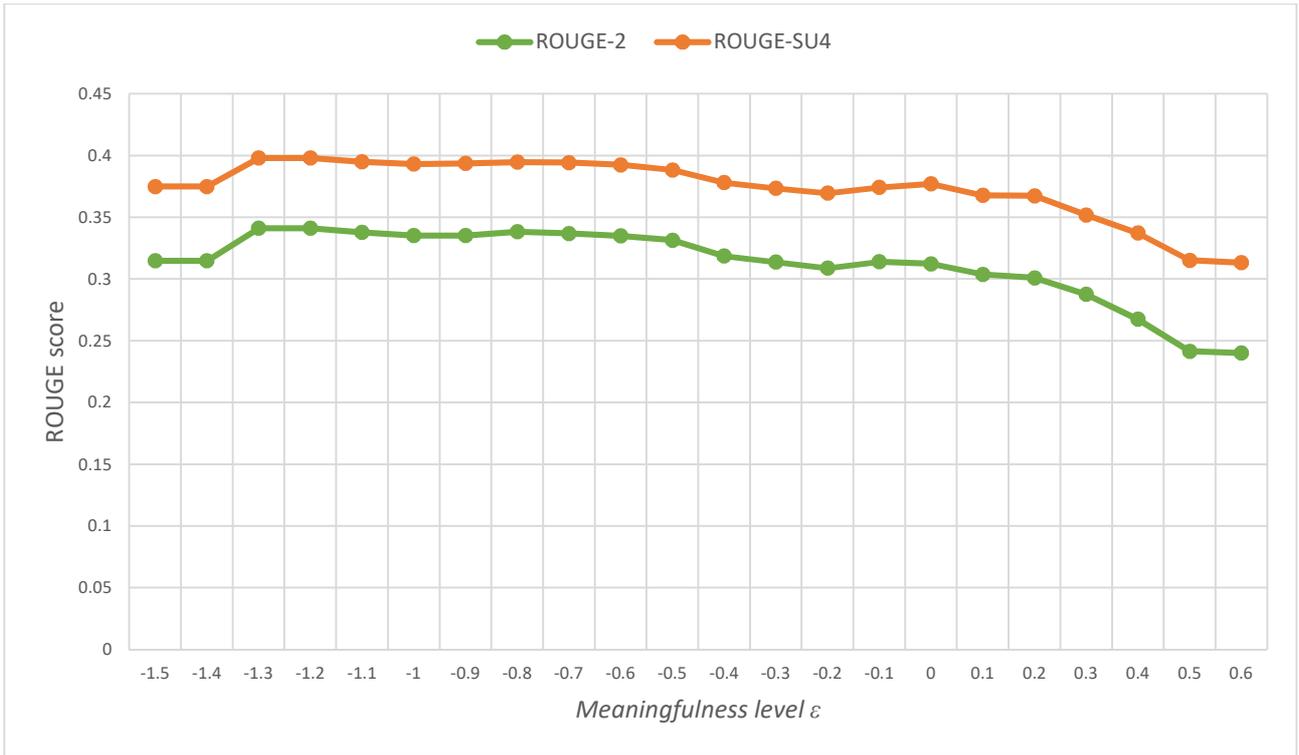

**Fig. 8.** ROUGE scores for the Bayesian summarizer using the fourth feature selection method and the different values of the meaningfulness threshold.

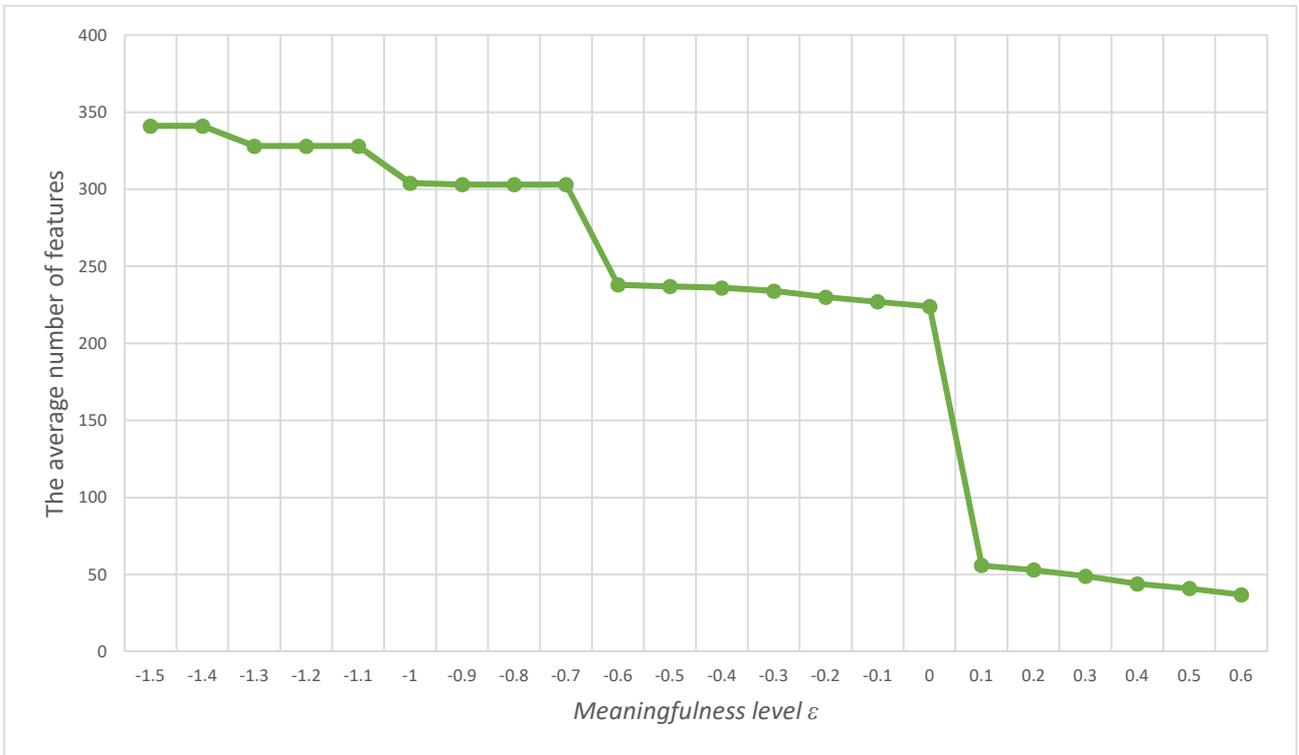

**Fig. 9.** The average number of features for the different values of the meaningfulness level in the fourth feature selection method.

In the fifth feature selection approach, we use frequent itemset mining to extract correlated concepts, i.e. frequent itemsets, and to select them as classification features. The itemset mining method employs a parameter $\varphi$ as the minimum support threshold to discover frequent itemsets. Fig. 10 shows The ROUGE scores obtained by the Bayesian summarizer using the fifth feature selection method and different values of the threshold $\varphi$ between [0.02, 0.23]. The threshold value of 0.09 reports the best scores (R-2: **0.3543** and R-SU4: **0.4094**). We choose this value as the optimal parameter for the subsequent experiments. Table 2 presents the average number of frequent itemsets selected as features for different values of the threshold $\varphi$ in the given range.

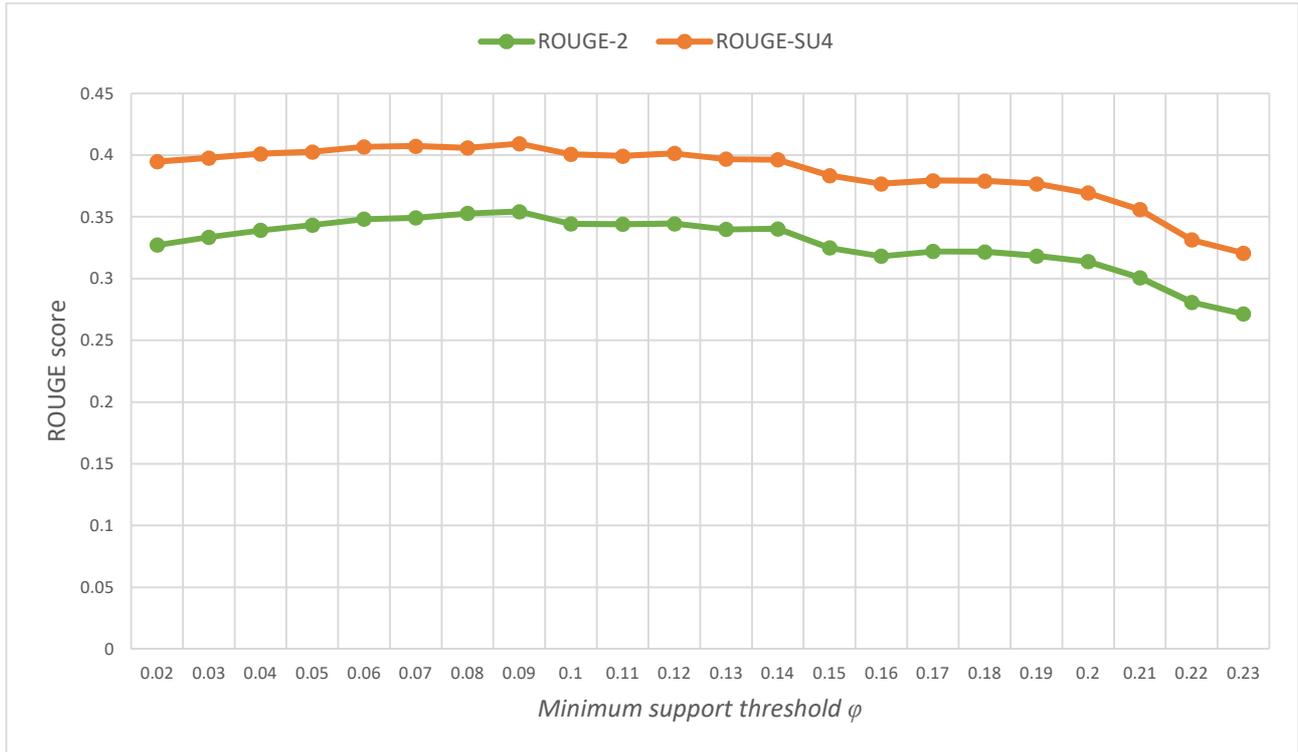

**Fig. 10.** ROUGE scores for the Bayesian summarizer using the fifth feature selection method and the different values of the minimum support threshold.

### 5.1.2. The impact of the coefficients

We mention in Section 3.4 that for the majority of documents, even the high-frequent concepts appear in less than 50% of the sentences of a document. We examine the most frequent concept of every document in the development corpus. The results show that only in nine documents there are concepts that appear in more than 50% of the sentences of the document. This means there are only nine documents containing at least one feature with a most probable value of True. For the other 91 documents, the most probable value of all the features is False. In average, the most frequent concept of a document in the development corpus appears in 36% of sentences in the corresponding document. We also investigate these statistics for the evaluation corpus. The results show that only in 33 documents there are concepts that appear in more than 50% of the sentences of the document. For the other 367 documents, the most frequent concept of each document appears in less than 50% of the sentences. In average, the most frequent concept of a document in the evaluation corpus appears in 35% of sentences in the corresponding document. Regarding these observations, as noted in Section 3.4, the coefficients help the summarizer to discriminate between the True and False values of features, leading to a more accurate sentence classification.

**Table 2.** The average number of features for different values of the minimum support threshold in the fifth feature selection method (extraction of correlated features by itemset mining).

| $\varphi$ | The average number of features | $\varphi$ | The average number of features |
|---|---|---|---|
| **0.02** | 2549 | **0.13** | 19 |
| **0.03** | 975 | **0.14** | 17 |
| **0.04** | 557 | **0.15** | 15 |
| **0.05** | 322 | **0.16** | 13 |
| **0.06** | 166 | **0.17** | 12 |
| **0.07** | 133 | **0.18** | 10 |
| **0.08** | 74 | **0.19** | 7 |
| **0.09** | 60 | **0.2** | 6 |
| **0.1** | 52 | **0.21** | 5 |
| **0.11** | 36 | **0.22** | 3 |
| **0.12** | 24 | **0.23** | 3 |

We perform a set of experiments to assess the impact of the coefficients $C_1$ and $C_2$ on the quality of the produced summaries. We evaluate the Bayesian summarizer using the best settings of all the five feature selection methods, with and without using the coefficients. Table 3 shows the ROUGE scores for these experiments. As can be seen, the summarizer reports higher scores when it utilizes the coefficients.

**Table 3.** ROUGE scores obtained by the Bayesian summarizer using the five different feature selection approaches, with and without using the coefficients.

|  | ROUGE-2 | | ROUGE-SU4 | |
|---|---|---|---|---|
|  | With the coefficients | Without the coefficients | With the coefficients | Without the coefficients |
| *First approach* | 0.3097 | 0.2762 | 0.3740 | 0.3435 |
| *Second approach* | 0.3246 | 0.2966 | 0.3857 | 0.3592 |
| *Third approach* ($\theta_3$) | 0.3346 | 0.3155 | 0.3932 | 0.3743 |
| *Fourth approach* ($\varepsilon=-1.2$) | 0.3411 | 0.2724 | 0.3980 | 0.3429 |
| *Fifth approach* ($\varphi=0.09$) | 0.3543 | 0.3391 | 0.4094 | 0.3919 |

### 5.1.3. The impact of the redundancy reduction method

Performing another set of preliminary experiments, we assess the impact of the redundancy reduction method on the quality of the produced summaries. Table 4 gives the ROUGE scores for the Bayesian summarizer when the five different feature selection approaches are used, with and without using the redundancy reduction method.

We compare the R-2 scores assigned to the Bayesian summarizer in two cases of using and not using the redundancy reduction method under different compression rates. Table 5 presents the percentage of improvement obtained using the redundancy reduction method for different compression rates. As can be seen, the percentage of improvement for the smaller compression rates is higher than for the greater rates.

**Table 4.** ROUGE scores obtained by the Bayesian summarizer using the five different feature selection approaches, with and without using the redundancy reduction method.

|  | *ROUGE-2* | | *ROUGE-SU4* | |
| --- | --- | --- | --- | --- |
|  | Redundancy reduction: Yes | Redundancy reduction: No | Redundancy reduction: Yes | Redundancy reduction: No |
| ***First approach*** | 0.3097 | 0.2861 | 0.3740 | 0.3589 |
| ***Second approach*** | 0.3246 | 0.3094 | 0.3857 | 0.3691 |
| ***Third approach ($\theta_3$)*** | 0.3346 | 0.3202 | 0.3932 | 0.3770 |
| ***Fourth approach ($\varepsilon=-1.2$)*** | 0.3411 | 0.3289 | 0.3980 | 0.3857 |
| ***Fifth approach ($\varphi=0.09$)*** | 0.3543 | 0.3415 | 0.4094 | 0.3903 |

**Table 5.** The percentage of improvement achieved by the Bayesian summarizer using the redundancy reduction method. The percentages have been computed for ROUGE-2 scores using the five different feature selection methods under different compression rates of 15%, 20%, 25%, 30%, and 35%.

|  | *Compression rate* | | | | |
| --- | --- | --- | --- | --- | --- |
|  | *15%* | *20%* | *25%* | *30%* | *35%* |
| ***First approach*** | 9.1 | 8.9 | 8.5 | 8.2 | 7.9 |
| ***Second approach*** | 5.9 | 5.6 | 5.2 | 4.9 | 4.7 |
| ***Third approach ($\theta_3$)*** | 5.5 | 5.1 | 4.9 | 4.5 | 4.4 |
| ***Fourth approach ($\varepsilon=-1.2$)*** | 4.8 | 4.5 | 4.1 | 3.7 | 3.5 |
| ***Fifth approach ($\varphi=0.09$)*** | 4.7 | 4.5 | 4.1 | 3.7 | 3.4 |

**Table 6.** The average number of concepts covered in the summaries produced by the Bayesian summarizer, using the five different feature selection approaches, with and without using the redundancy reduction method.

|  | *With redundancy reduction* | *Without redundancy reduction* |
|---|---|---|
| *First approach* | 330 | 297 |
| *Second approach* | 336 | 312 |
| *Third approach* ($\theta_3$) | 302 | 272 |
| *Fourth approach* ($\varepsilon=-1.2$) | 332 | 308 |
| *Fifth approach* ($\varphi=0.09$) | 303 | 284 |

We also assess the average number of concepts covered in the produced summaries by the Bayesian summarizer, with and without using the redundancy reduction method. Table 6 shows these results when the summarizer uses the five different feature selection methods. As the results show, the summaries cover more concepts in average when the summarizer benefits from the redundancy reduction method.

### 5.2. Evaluation results

Comparing the Bayesian summarizer with the other methods, we evaluate the performance of our method for biomedical text summarization. Table 7 shows the ROUGE scores obtained by the Bayesian summarizer and the comparison methods. We evaluate the Bayesian summarizer all the five feature selection approaches. In order to test the statistical significance of the results, we use a Wilcoxon signed-rank test with a 95% confidence interval.

According to the results, when the Bayesian summarizer uses the fourth and fifth feature selection methods, it significantly outperforms all the comparison methods in terms of all the reported ROUGE scores ($p < 0.05$). Using the third feature selection method, the Bayesian summarizer significantly performs better than BioChain, the domain-independent, and the baseline methods in terms of all the ROUGE scores ($p < 0.05$). In comparison with ChainFreq and FreqDist, the results report a significant improvement for all the scores except for the R-W-1.2 ($p > 0.05$). When we run the Bayesian summarizer with the second feature selection method, it significantly performs better than the domain-independent and baseline summarizers ($p < 0.05$). Among the biomedical summarization methods, its improvement is significant for all the ROUGE scores compared to BioChain, and its improvement is significant only for R-2 score with respect to FreqDist ($p < 0.05$). Although it obtains better scores than ChainFreq, its improvement is not significant for all the scores ($p > 0.05$). Eventually, the Bayesian summarizer significantly performs better than the domain independent competitors and baselines in terms of all the scores when it uses the first feature selection method ($p < 0.05$). Compared to BioChain, it improves all the scores, but the improvement is only significant for the R-1.

Table 8 gives the average, minimum, and maximum number of features selected by the five feature selection methods for documents in the evaluation corpus.

**Table 7.** ROUGE scores obtained by the Bayesian summarizer and the comparison methods. The best result for each ROUGE score is shown in bold type. The summarizers are sorted based on the decreasing order of their ROUGE-2 scores.

|  | *ROUGE-1* | *ROUGE-2* | *ROUGE-W-1.2* | *ROUGE-SU4* |
|---|---|---|---|---|
| *Bayesian summarizer- Fifth approach ($\varphi=0.09$)* | **0.7886** | **0.3529** | **0.1113** | **0.4104** |
| *Bayesian summarizer- Fourth approach ($\varepsilon=-1.2$)* | 0.7760 | 0.3442 | 0.0975 | 0.4019 |
| *Bayesian summarizer- Third approach ($\theta_3$)* | 0.7634 | 0.3351 | 0.0862 | 0.3961 |
| *Bayesian summarizer- Second approach* | 0.7549 | 0.3263 | 0.0808 | 0.3872 |
| *ChainFreq* | 0.7507 | 0.3182 | 0.0798 | 0.3791 |
| *FreqDist* | 0.7498 | 0.3121 | 0.0789 | 0.3782 |
| *Bayesian summarizer- First approach* | 0.7496 | 0.3114 | 0.0785 | 0.3759 |
| *BioChain* | 0.7378 | 0.3080 | 0.0746 | 0.3691 |
| *SUMMA* | 0.7006 | 0.2865 | 0.0714 | 0.3418 |
| *SweSum* | 0.6997 | 0.2817 | 0.0703 | 0.3386 |
| *Lead baseline* | 0.6351 | 0.2459 | 0.0681 | 0.3035 |
| *AutoSummarize* | 0.6158 | 0.2407 | 0.0656 | 0.2948 |
| *Random baseline* | 0.5602 | 0.2243 | 0.0615 | 0.2711 |

**Table 8.** The average, minimum, and maximum number of features for the documents of the evaluation corpus using the five different feature selection approaches.

|  | *Average* | *Minimum* | *Maximum* |
|---|---|---|---|
| *Bayesian summarizer- First approach* | 980 | 399 | 1625 |
| *Bayesian summarizer- Second approach* | 519 | 175 | 925 |
| *Bayesian summarizer- Third approach ($\theta_3$)* | 22 | 6 | 35 |
| *Bayesian summarizer- Fourth approach ($\varepsilon=-1.2$)* | 344 | 117 | 620 |
| *Bayesian summarizer- Fifth approach ($\varphi=0.09$)* | 61 | 7 | 399 |

# 6. Discussion

## 6.1. Feature selection and parameterization

As reported in Table 1, when the summarizer uses the third feature selection method with the threshold $\theta_3$, the scores are higher than for the other two threshold values. For a given document, the value of the $\theta_1$ is always less than for the other two thresholds, and the value of $\theta_3$ is always greater than for the other ones. For the preliminary experiments evaluation corpus, the average number of selected features for a document using the thresholds $\theta_1$, $\theta_2$, and $\theta_3$ is equal to 108, 35, and 19, respectively. This shows that when we use the $\theta_1$ as the threshold value, the number of selected features is relatively high. In this case, only some of the features indicate to essential concepts, and the summarizer can be misled by numerous unimportant features. On the other hand, when we use the $\theta_2$ and $\theta_3$, the number of selected features decreases to almost less than one third. This reduction in the number of features helps the summarizer to decide more accurately, considering the features which point to important concepts indeed.

As Fig. 8 shows, when the summarizer uses the fourth feature selection method, the best scores are reported for the both meaningfulness level of −1.3 and −1.2. In this case, the average number of features is 328. As showed in Fig. 9, for the threshold values greater than zero, the average number of features falls to nearly 50. This rapid decrease happens because for the majority of concepts the meaningfulness value is zero. As the number of features decreases rapidly, the quality of summarization also decreases, because many features which could help the summarizer to perform more accurately are no longer available. The results show that the meaningfulness threshold of −1.3 discards many features which can be considered as redundant ones. When we assign threshold values smaller than -1.3, redundant features decrease the performance of the summarizer.

There are other measures, such as Inverse Document Frequency (IDF), Inverse Sentence Frequency (ISF), and Inverse Term Frequency (ITF) that are widely adopted in text mining research [58]. Such measures may seem to be functionally similar to the meaningfulness measure defined by the Helmholtz principle. However, by studying the theoretical and practical functions of the measures, we found the meaningfulness more useful than others for selecting important concepts in the Bayesian summarizer for several reasons. First, the IDF, ISF, and ITF measures are generally defined for a corpus of documents, but the Bayesian summarizer analyzes one document at a time, and the definition of such measures may not make sense for single documents. Second, if we define the ISF or ITF weights in a document, concepts appearing in fewer sentences are assigned more discriminative power. Such weighting scheme could not be useful in our context. On the other hand, the meaningfulness measure has been adopted successfully in single-document text analysis. Furthermore, as the example in Fig. 4 shows, the weights assigned by the Helmholtz principle do not have any obvious relation to the frequency of concepts. According to the formulas in Section 3.2.4, the meaningfulness value depends on multiple factors, such as the frequency of concepts within each paragraph and within the whole document, the length of each paragraph, and the length of the document.

As Fig. 10 shows, when the summarizer uses the fifth feature selection method, the best scores are reported for the minimum support threshold of 0.09. For this value, the average number of features (itemsets) is 60. For the smaller thresholds which produce more number of features, the ROUGE scores decrease slightly. This shows that more number of features could not significantly reduce the accuracy of the summarizer in this feature selection method. Although the numerous features mislead the summarizer to some extent, it still benefits from the knowledge about the correlated concepts provided by the itemsets. When the threshold $\varphi$ tends to be greater than 0.09, the average number of features decreases, and the performance of the summarizer also decreases. Particularly, when the threshold is greater than 0.14, the average number of features drops to less than 17 and the ROUGE scores are reduced considerably. This shows that when the number of features decreases in fifth feature selection method, the knowledge of the summarizer about important and correlated concepts is inadequate. It decides according to a limited number of high-supporting itemsets whereas there are a lot of useful itemsets discarded by an extreme threshold.

**6.2. The coefficients $C_1$ and $C_2$**

As can be seen in Table 3, for all the five feature selection methods, the Bayesian summarizer obtains better ROUGE scores when it uses the coefficients. Since for the majority of documents even the most frequent features do not appear in more than 50% of the sentences of a document, the most probable value for almost every feature is *False*. When the method does not use the coefficients, it decides based on the most probable values of the features. In this case, the summarizer needs additional knowledge about the features and their importance to decide more accurately. Using the coefficients, when the value of a feature is True in a sentence, the probability of selecting the sentence for the summary increases in proportion to the occurrence of the feature. Moreover, the probability of not selecting the sentence decreases. On the other hand, when the value of a feature is False in a sentence, the probability of selecting the sentence for the summary decreases in proportion to the occurrence of the feature. Moreover, the probability of not selecting the sentence increases. Adopting this strategy, the summarizer can discriminate between the presence and absence of important concepts. As a result, the summarizer performs better and reports higher scores.

**6.3. The redundancy reduction method**

As Table 4 shows, when the redundancy reduction method takes part in the summarization method, the summarizer reports better scores. With the help of the redundancy reduction method, the summarizer gives sentences containing low-frequency concepts a higher chance to be included in the summary. Therefore, the summary can cover more number of concepts while it still conveys important concepts and subtopics. When summaries cover more relevant information, their informativeness increases. As a results, the summarizer obtains higher scores.

Table 5 suggests that the percentage of improvement for the smaller compression rates is higher than for the greater rates. This happens because for smaller compression rates, fewer sentences must be selected for the summary. Hence, when we use the redundancy reduction method and the summarizer selects new sentences containing new information, the scores report a more impressive improvement. On the other hand, for greater compression rates, the summarizer selects more sentences. In this way, the summary automatically includes new information. The summary presents the new information along with some potentially redundant sentences which were selected earlier or would be selected later. The redundancy reduction method may discard these redundant sentences and select new relevant information. Therefore, the performance of the summarizer may be improved. However, this improvement for greater compression rates is less than for smaller rates, because new relevant information is automatically included in the summary by the greater compression rates.

As Table 6 shows, using the redundancy method, the summaries cover more concepts in average. In addition, as can be seen in Table 4, the usage of the redundancy reduction method leads to an increase in the scores for all the feature selection methods. However, when we compare each pair of feature selection methods, the greater average number of concepts covered in the produced summaries does not necessarily lead to better summarization performance. For example, in both cases of using and not using the redundancy reduction method, the average number of concepts covered in the summaries for the fifth feature selection method is less than the average for the first, second and fourth methods. Nevertheless, the summarizer obtains the best ROUGE scores using the fifth method. This suggests that with the use of an appropriate feature selection method, the summaries convey more informative content even if they cover fewer concepts. These results demonstrate that both the redundancy reduction method and an appropriate feature selection method are essential to enhance the performance of the summarizer. The lack of each one has a negative impact on the quality of produced summaries.

### 6.4. Comparison with other summarizers

As Table 7 shows, when the Bayesian summarizer utilizes the first feature selection method, it performs better than the domain-independent competitors. This suggests that using concepts as classification features in our method can be a better approach compared to the summarizers which employ word frequency methods, positional features, and term similarity features. Our summarizer performs slightly worse than the two biomedical summarizers, i.e. FreqDist and ChainFreq, when it considers all extracted concepts as features. In this case, it seem that potentially redundant and unrelated concepts negatively affect the quality of produced summaries.

Comparing the results of the first and second feature selection methods, we observe that when generic and potentially redundant concepts are discarded, the summarizer can decide more accurately and obtains higher scores. Looking at the number of features selected by the first and second methods in Table 8, it seems that almost a half of concepts extracted from a document can be considered as unnecessary. Removing redundant concepts from classification features, we observe a slight increase in the performance of the summarizer. The results of the second feature selection method show that, with respect to BioChain, considering the distribution of non-generic concepts along with the redundancy reduction method improves the performance of biomedical summarization. With respect to FreqDist and ChainFreq, we still need to make more refinement in our feature selection strategy to increase the quality of produced summaries.

The third feature selection method employs a ranking and filtering strategy. As the results show, the use of all extracted concepts to constructing a frequency distribution model, i.e. FreqDist, can be outperformed by the Bayesian modeling in combination with an optimized feature selection based on the filtering method. Although ChainFreq does not use all extracted concepts and utilizes BioChain as a filtering method for its hybrid method, the Bayesian summarizer obtains relatively better scores using an appropriate threshold for filtering the features. When we employ the third method for feature selection, the maximum, minimum and average number of features for a document in the evaluation corpus is 35, 6, and 22 respectively. As the numbers in Table 8 show, the third method considerably reduces the average number of features compared to the second method, from 519 to 22. This reduction helps the summarizer to decide more accurately. Regarding the results of the second method, this filtering strategy leads to a slight increase in the performance of summarization.

The results of the fourth feature selection method show that the meaningfulness is a better measure than the frequency for feature selection in the Bayesian summarizer. As Table 8 shows, the fourth method produces a large number of features compared to the third method, 344 versus 22 for the average number of features. However, the fourth method yields better summarization performance. This suggests that the meaningfulness can be considered as a more efficient measure to remove unimportant concepts in the Bayesian summarizer. This measure provides the summarizer with a set of indeed important concepts to decide more optimally. However, some of concepts selected as features may not be considered as a frequent concept. When the summarizer utilizes this measure, it still makes use of the frequency of concepts in the form of the coefficients. In fact, the summarizer combines information about the meaningfulness measure and the frequency of concepts. Using this approach, the Bayesian summarizer obtains higher scores than all the biomedical competitors.

As Table 7 shows, the Bayesian summarizer reports the highest scores when it utilizes the fifth feature selection method and uses frequent itemsets as features. Using this method, the summarizer implicitly takes into account correlations and appearing patterns existing among concepts. The results show that this feature selection strategy and the Bayesian modeling yield better summarization quality than the biomedical summarizers relying on the frequency of single concepts. According to the results, the fifth feature selection method performs slightly better than the fourth method. This suggests that the Bayesian summarizer can utilize either information about correlated concepts or the meaningfulness measure as two useful feature selection approaches to improve the performance of summarization. Comparing the results of different feature selection methods, we observe that frequent itemsets can be more useful than the frequency of single concepts in the Bayesian summarizer.

According to the results, our Bayesian summarizer significantly outperforms the domain-independent comparison methods, i.e. SUMMA, SweSum, and AutoSummarize, in biomedical text summarization. These methods utilize statistical, similarity-based, and word frequency features for sentence selection. The results show that these term-based methods cannot be considered as useful summarizers for biomedical text. On the other hand, using domain knowledge and efficient feature selection methods, the Bayesian summarizer can perform more efficiently than the comparison methods.

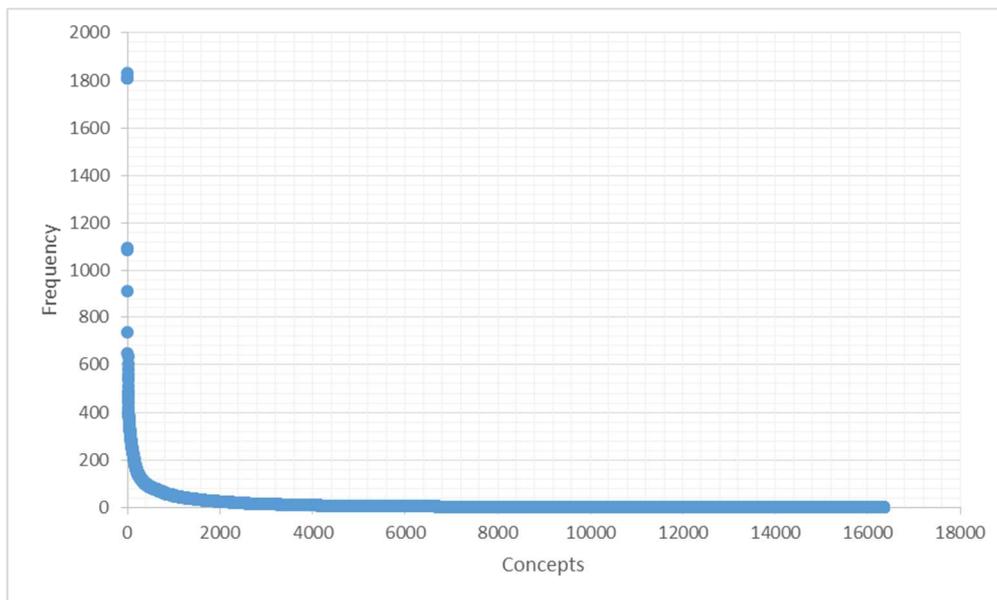

**Fig. 11.** The frequency distribution of concepts within the full-text papers in the evaluation corpus.

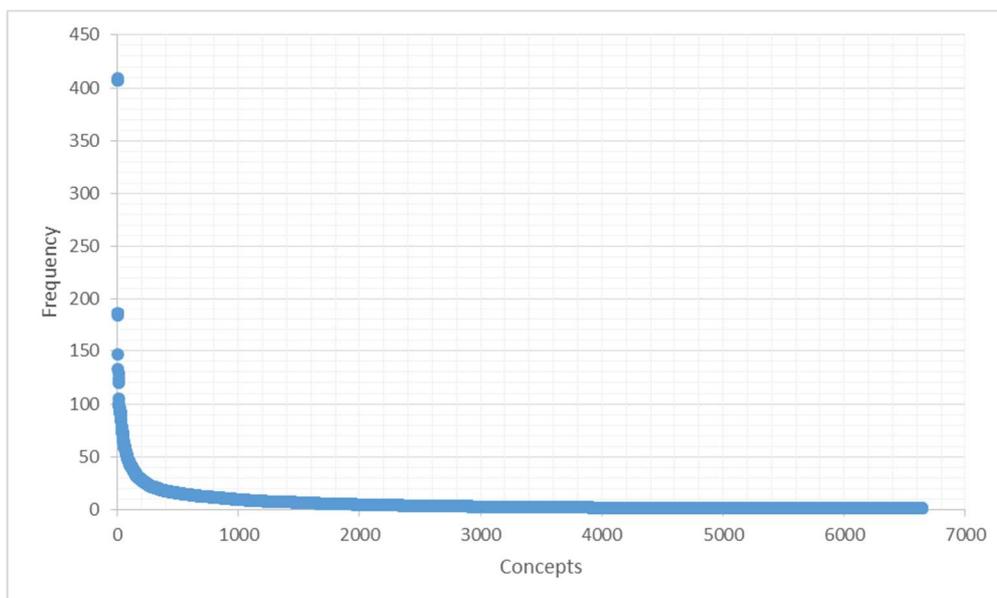

**Fig. 12.** The frequency distribution of concepts within the abstracts in the evaluation corpus.

### 6.5. Justification of the basic assumption

As we explained in Section 3.4, the Bayesian summarizer estimates the posterior probability of selecting and not selecting sentences for the summary based on the prior probability of concepts within the input document. In other words, the summarizer assumes that the distribution of important concepts within the summary should be similar to the original text. This assumption has been justified by Reeve et al. [10]. They selected a corpus of 24 biomedical full-text papers and used the abstracts of the papers as the ideal summaries. They constructed two frequency distribution models from the concepts of the full-text papers and the abstracts and showed that these two frequency distribution models follow Zipfian distribution. Regarding this observation, they suggested that a full-text paper and a version of its ideal summary (abstract) have the same frequency distribution form. We perform a similar experiment using a larger corpus. We extract concepts from the full-texts and the abstracts of our evaluation corpus containing 400 biomedical papers. Fig. 11 shows the frequency distribution of 16,353 concepts within the full-text papers, and Fig. 12 shows the frequency distribution of 6,645 concepts within the abstracts. As can be observed from Fig. 11 and Fig. 12, both the full-texts and the abstracts (considered as the ideal summaries) follow Zipfian distribution. This observation justifies the basic assumption of our proposed biomedical text summarization method that uses the distribution of concepts within the document to estimate the posterior probability of the sentences for inclusion in the summary.

## 7. Conclusion

In this paper, we propose a biomedical text summarization method using a Bayesian classification approach. The method classifies the sentences of the input document as summary and non-summary according to the distribution of important concepts within the text. We introduce different feature selection approaches to identify the important concepts of the document and using them as classification features. The summarizer uses two coefficients in probability estimation to discriminate between the presence and absence of important concepts. It also employs a simple redundancy reduction method to reduce the potential redundancy in the summary. Conducting a set of preliminary experiments on a development corpus containing 100 biomedical papers, we tune the parameters of the system and assess the efficiency of the coefficients and the redundancy reduction method. The results show that the coefficients and the redundancy reduction method have a positive impact on the quality of produced summaries, leading to an improvement in the performance of the summarizer.

We evaluate the performance of the Bayesian summarization method in comparison with the other biomedical summarizers relying on the frequency of concepts, domain-independent summarizers, and baseline methods using an evaluation corpus of 400 biomedical papers. The results show that when the Bayesian summarizer utilizes the meaningfulness measure rather than the frequency of single concepts for selecting features, it outperforms the other summarizers. Moreover, when the summarizer employs itemset mining and uses correlated concepts as classification features, it significantly performs better than the comparison methods. Summing up the results, we can draw the following conclusions that answer to the questions raised in Section 1:

- A Bayesian classification method can be utilized for the probability distribution modeling of concept-based biomedical text summarization. An efficient feature selection method is required to enhance the accuracy of the classification method.
- The summarizer should not consider all the extracted concepts from the input document. There many redundant concepts which may have a negative impact on the accuracy of the model and can be discarded by the summarizer.
- The meaningfulness measure defined by the Helmholtz principle can be a useful criterion, rather than the frequency, to identify important concepts and use them as classification features.
- Using itemset mining to discover correlated concepts and incorporating these correlations into the feature selection phase provide the summarizer with a more accurate model, leading to an increase in the obtained scores.

In our future research we intend to concentrate on extending our Bayesian biomedical summarizer to deal with multi-document and query-focused summarization. To do so, a more complicated redundancy reduction method should be studied. It seems that there is much more room for studying the Helmholtz principle from the Gestalt theory in the context of concept-based summarization. Balinsky et al. [59] have modeled document sentences as a small world network using the Helmholtz principle and investigated some applications such as text summarization. Future work can involve exploring this type of modeling for concept-based biomedical text summarization. The study of using other discriminative classifiers in this type of summarization can be considered as another potential topic for future research.

**Conflict of interest**

The authors declare that they have no conflict of interest.